\documentclass[runningheads]{llncs}
\usepackage{graphicx}
\usepackage{tikz}
\usepackage{comment} 
\usepackage{amsmath,amssymb} % define this before the line numbering.
\usepackage{color}

\usepackage{algorithm}
\usepackage{algorithmic}
\usepackage{multirow}
\usepackage{tabularx}
\usepackage{booktabs}

\graphicspath{{figures/}} %画像を別ディレクトリにまとめる

\newcommand{\Tref}[1]{Table~\ref{#1}}
\newcommand{\Eref}[1]{Eq.~(\ref{#1})}
\newcommand{\Fref}[1]{Fig.~\ref{#1}}
\newcommand{\Aref}[1]{Algorithm~\ref{#1}}
\newcommand{\Sref}[1]{Section~\ref{#1}}
\newcommand{\Cref}[1]{Chapter~\ref{#1}}

\newcolumntype{L}[1]{>{\raggedright\arraybackslash}p{#1}}
\newcolumntype{C}[1]{>{\centering\arraybackslash}p{#1}}
\newcolumntype{R}[1]{>{\raggedleft\arraybackslash}p{#1}}

\begin{document}
% \linenumbers
\pagestyle{headings}
\mainmatter
\def\ECCVSubNumber{100}  % Insert your submission number here

\title{Multi-Task Curriculum Framework for \\Open-Set Semi-Supervised Learning} % Replace with your title

\begin{comment}
\titlerunning{ECCV-20 submission ID \ECCVSubNumber} 
\authorrunning{ECCV-20 submission ID \ECCVSubNumber} 
\author{Anonymous ECCV submission}
\institute{Paper ID \ECCVSubNumber}
\end{comment}
%******************

%\begin{comment}
\titlerunning{Multi-Task Curriculum Framework for Open-Set SSL}
\author{Qing Yu\inst{1} \and
Daiki Ikami\inst{1,2} \and
Go Irie\inst{2} \and
Kiyoharu Aizawa\inst{1}}
\authorrunning{Q. Yu et al.}
\institute{The University of Tokyo, Japan \\
\email{\{yu,ikami,aizawa\}@hal.t.u-tokyo.ac.jp} \and
NTT Corporation, Japan\\
\email{goirie@ieee.org}}
%\end{comment}
%******************
\maketitle

\begin{abstract}
Semi-supervised learning (SSL) has been proposed to leverage unlabeled data for training powerful models when only limited labeled data is available. While existing SSL methods assume that samples in the labeled and unlabeled data share the classes of their samples, we address a more complex novel scenario named open-set SSL, where out-of-distribution (OOD) samples are contained in unlabeled data. Instead of training an OOD detector and SSL separately, we propose a multi-task curriculum learning framework. First, to detect the OOD samples in unlabeled data, we estimate the probability of the sample belonging to OOD. We use a joint optimization framework, which updates the network parameters and the OOD score alternately. Simultaneously, to achieve high performance on the classification of in-distribution (ID) data, we select ID samples in unlabeled data having small OOD scores, and use these data with labeled data for training the deep neural networks to classify ID samples in a semi-supervised manner. We conduct several experiments, and our method achieves state-of-the-art results by successfully eliminating the effect of OOD samples.

\end{abstract}

\section{Introduction}
After several breakthroughs in deep learning methods, deep neural networks (DNNs) have achieved impressive results and even outperformed humans on various machine perception tasks such as image classification \cite{he2016deep}\cite{tan2019efficientnet}, face recognition \cite{Liu_2017_CVPR}, and natural language processing \cite{devlin2018bert} with large-scale, annotated training samples. However, creating these large datasets is typically time-consuming and expensive.

To solve this problem, semi-supervised learning (SSL) is proposed to leverage unlabeled data to improve the performance of a model when only limited labeled data is available. SSL is able to train large, powerful models when labeling data is expensive or inconvenient. There is a diverse collection of approaches to SSL. For example, one approach is consistency regularization \cite{sajjadi2016regularization}\cite{laine2016temporal}\cite{tarvainen2017mean}, which encourages a model to produce the same prediction when the input is perturbed. Another approach, entropy minimization \cite{grandvalet2005semi}, encourages the model to produce high-confidence predictions. The recent state-of-the-art method, MixMatch \cite{berthelot2019mixmatch}, combines the aforementioned techniques in a unified loss function and achieves strong performance on a variety of image classification benchmarks.

These existing SSL methods assume that the labeled and unlabeled data have the same distribution, meaning that they share the classes of their samples, and there is no outlier sample in unlabeled data. However, in the real world, it is hard to ensure that the unlabeled data does not contain any out-of-distribution (OOD) sample that is drawn from different distributions. Oliver et al. \cite{oliver2018realistic} have shown that adding unlabeled data from a mismatched set of classes can actually damage the performance of SSL.

\begin{figure}[t]
    \centering
    \includegraphics[width=0.9\textwidth]{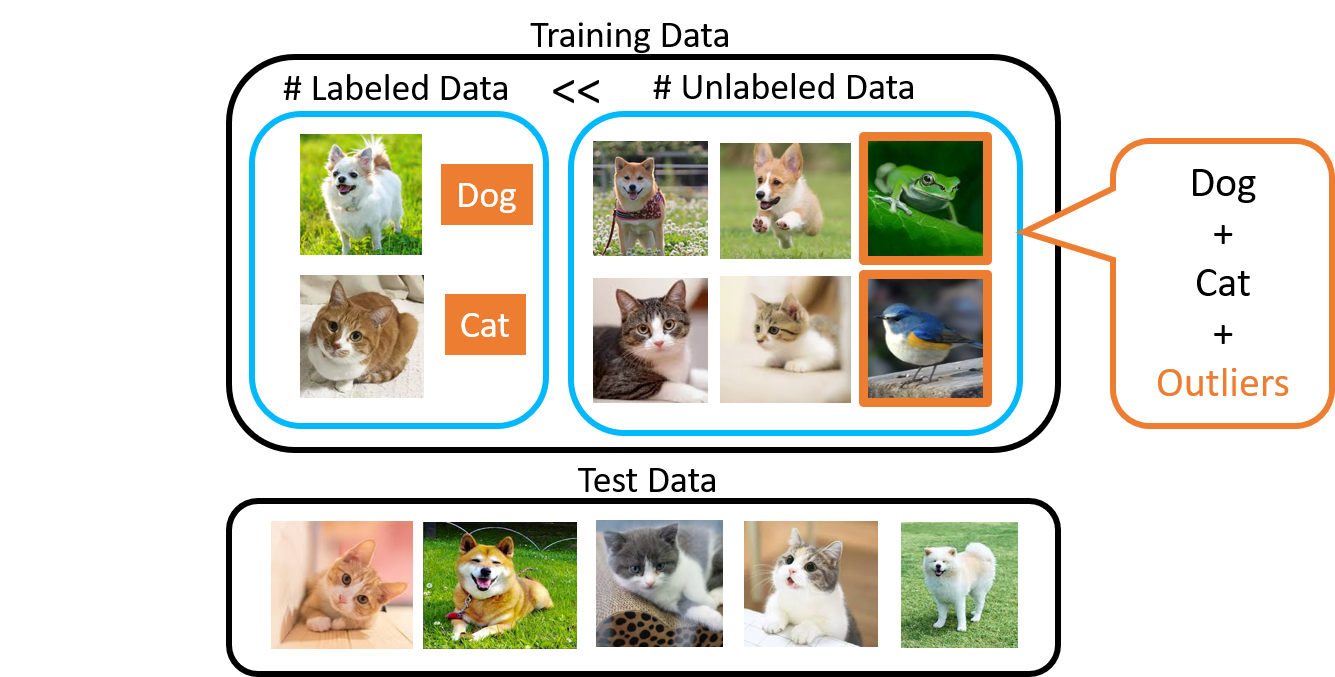}
    \caption{Problem setting of open-set SSL. Outliers, which do not belong to any class of labeled data, exist in the unlabeled data.}
    \label{fig:setting}
\end{figure}

Hence, we consider a new, realistic setting called ``Open-Set Semi-supervised Learning'', as shown in \Fref{fig:setting}. Outliers, which do not belong to the classes of labeled data, exist in the unlabeled data, and the model should be trained on labeled and unlabeled data by eliminating the effect of these outliers. To the best of our knowledge, our study is the first to tackle the problem of open-set SSL.

Although there are many algorithms for detecting OOD samples \cite{hendrycks2016baseline}\cite{lee2017training}\cite{liang2017enhancing}\cite{vyas132018out}, these methods are trained on a large number of labeled in-distribution (ID) samples with class labels. In the setting of SSL, the number of labeled data is very limited. Hence, the previous methods cannot achieve high performance of detection and are not suitable for open-set SSL. Therefore, we propose a method that uses multi-task curriculum learning, which is a multi-task framework aiming to solve OOD detection and SSL simultaneously. 

First, we detect OOD samples in the unlabeled data. We propose a new OOD detection method by a joint optimization framework, which can utilize the unlabeled data containing OOD data in the process of training an OOD detector. We train the network to estimate the probability of the sample belonging to OOD. At the beginning of training, we treat all unlabeled samples as OOD and all labeled samples as ID by assigning an initial OOD score to each sample (0 for labeled data and 1 for unlabeled data). Next, we train the model to classify the sample as OOD or ID. Since unlabeled data also contains a reasonable amount of ID samples, treating all unlabeled samples as OOD samples would result in incorrect label assignments. Inspired by a solution of the noisy label problem \cite{tanaka2018joint}, we update the network parameters and the OOD scores alternately as a joint optimization to clean the noisy OOD scores of unlabeled samples, which ranges from 0 to 1.

At the same time, while training the network for OOD detection, we also train the network to classify ID samples correctly, which forms multi-task learning. Since ID samples in the unlabeled data are expected to have smaller OOD scores than the real OOD samples, we use curriculum learning that excludes the samples with higher OOD scores in unlabeled data. Then we combine remaining ID unlabeled samples with labeled data for training the CNN to classify ID samples correctly by any SSL method, where MixMatch \cite{berthelot2019mixmatch} is used in this paper.

We evaluate our method on a diverse set of open-set SSL settings. In many settings, our method outperforms existing methods by a large margin. We summarize the contributions of this paper as the following:
\begin{itemize}
    \item We propose a novel experimental setting and training methodology for open-set SSL.

    \item We propose a multi-task curriculum learning framework that detects OOD samples by alternate optimization and classifies ID samples by applying SSL according to the results of OOD detection.

    \item We evaluate our method across several open-set SSL tasks and outperforms state-of-the-art by a considerable margin. Our approach successfully eliminates the effect of OOD samples in the unlabeled data.
\end{itemize}

\section{Related Work}
\label{sec:related}
At present, there are several different methods of SSL and OOD detection. We will briefly explain some important studies in this section.
\subsection{Semi-supervised Learning}
Although there are many studies on SSL techniques, such as transductive models \cite{joachims1999transductive}\cite{joachims2003transductive}, graph-based methods \cite{zhu2003semi} and generative modeling \cite{kingma2014semi}\cite{pu2016variational}\cite{salimans2016improved}, we focus mainly on the recent state-of-the-art methods, based on consistency regularization \cite{laine2016temporal}\cite{sajjadi2016regularization}\cite{tarvainen2017mean}.

In general supervised learning, data augmentation is a common regularization technique. In image classification, it is common to add some noise to an input image to change the pixel values of an image but keep its label \cite{cubuk2019autoaugment}, which means data augmentation is able to artificially increase the size of a training set by generating new modified data.

Consistency regularization is a method that applies data augmentation to SSL. It imposes a constraint in the form of regularization so that the classification result of each unlabeled sample does not change before and after augmentation.

In the simplest method, Laine and Aila \cite{laine2016temporal} proposed $\pi$-model, which applies two different stochastic augmentations to an unlabeled data to generate two inputs and minimize the distance of the two network outputs of these two inputs.

Mean Teacher \cite{tarvainen2017mean} used an exponential moving average of network parameter values to generate a more stable output on one of the two inputs, instead of generating two outputs for the two inputs by the same network, to improve the effectiveness of their method.

The state-of-the-art method for SSL is MixMatch \cite{berthelot2019mixmatch}, which works by guessing low-entropy labels for data-augmented unlabeled examples and mixing labeled and unlabeled data using MixUp \cite{zhang2017mixup}. MixMatch \cite{berthelot2019mixmatch} then uses $\pi$-model to train a model using the mixed labeled and unlabeled data. We refer the reader to their paper \cite{berthelot2019mixmatch} for further details.

However, these methods assume that the labeled and unlabeled data share the classes of their samples. When some OOD samples are contained in the unlabeled data, Oliver et al. \cite{oliver2018realistic} showed that the existing methods achieved bad performance, which is even lower than the performance of supervised learning trained only by limited labeled data in some cases. The motivation of this research is to solve this problem. 

\subsection{Out-of-distribution detection}

There are also some methods for OOD detection. In the simplest method, Hendrycks \& Gimpel \cite{hendrycks2016baseline} used the predicted softmax class probability to detect OOD samples. They observed that the prediction probability of incorrect and OOD samples tends to be lower than that of the correct samples. However, they also found that a pre-trained neural network can still classify some OOD samples overconfidently, which limits its performance.

To improve the effectiveness of Hendrycks \& Gimpel's method \cite{hendrycks2016baseline}, Liang et al. \cite{liang2017enhancing} applied temperature scaling and input preprocessing to detect OOD samples, called Out-of-DIstribution detector for Neural networks (ODIN). They found that the difference between the largest logit (the outputs of which are not normalized by softmax) and the remaining logits is larger for ID samples than for OOD samples if the logits are scaled by a large constant (temperature scaling). They showed that the separation of the softmax scores between the ID and OOD samples could be increased by temperature scaling. They also found that the addition of small perturbations to the input (through the loss gradient) increases the maximum predicted softmax score. As a result, the ID samples show a greater increase in score than the OOD samples. Using these techniques, their method outperformed the baseline method \cite{hendrycks2016baseline}.

In another method using the predicted probability of the network, Bendale \& Boult \cite{bendale2016towards} calculated the score for an unknown class by taking the weighted average of all other classes obtained from a Weibull distribution, named openMax layer.

However, all the methods described earlier need a large number of labeled ID samples to achieve stable results and they are unable to utilize any unlabeled data. In open-set SSL, the number of labeled ID samples is small but we have access to a huge amount of unlabeled data containing some OOD samples. Our method aims at training a model to not only detect OOD samples with limited labeled and plenty of unlabeled data, but also achieve high recognition performance on the classification of ID samples.

\section{Method}
\subsection{Problem Statement}
We assume that an ID image-label pair, $\{\boldsymbol{x}_{l},y_{l}\}$, drawn from a set of labeled ID images $\{X_{l}, Y_{l}\}$ ,  as well as an unlabeled image, $\boldsymbol{x}_{ul}$, drawn from a set of unlabeled images $X_{ul}$, is accessible. The labeled ID sample $\{\boldsymbol{x}_{l},y_{l}\}$ can be classified into one of $K$ classes denoted by $\{c_1, \dots, c_K\}$, meaning that $y_l \in \{c_1, \dots, c_K\}$. Besides ID samples, outliers (OOD samples) also exist in the unlabeled data, which signifies that the true class of some unlabeled data $\boldsymbol{x}_{ul}$ is not in $\{c_1, \dots, c_K\}$.

The goal of our method is to train a model that can correctly classify ID samples into $\{c_1, \dots, c_K\}$ on a combination of labeled ID samples and unlabeled samples under semi-supervised setting. Our technique achieves this by distinguishing whether the image $\boldsymbol{x}_{ul}$ is from in-distribution to eliminate the negative effect of OOD samples during the training.

\subsection{Overall Concept}
The most challenging part of this task is the detection of OOD samples when only limited labeled ID samples are available. As mentioned in \Sref{sec:related}, traditional OOD detection methods \cite{hendrycks2016baseline,liang2017enhancing} assumed that there is a large number of labeled ID samples for training the recognition model and these methods did not utilize unlabeled data in training. Thus, these methods cannot achieve high performance OOD detection in SSL.

We propose a multi-task curriculum learning framework for open-set SSL, which aims to solve OOD detection and SSL simultaneously as a multi-task framework.

Since the number of labeled ID samples is limited in SSL, we use a joint optimization framework inspired by \cite{tanaka2018joint}, which updates DNN parameters and estimates the probability of the sample belonging to OOD alternately. First, we assign an initial pseudo label representing the probability of the sample belonging to OOD, named OOD score, to all the data. For labeled samples, since they are ID, we initialize the OOD scores as 0 and for unlabeled samples, we initialize the OOD scores as 1. Since some amount of ID samples are present in unlabeled data, we can consider the binary classification of OOD as a noisy label problem. \cite{tanaka2018joint} showed that a DNN trained on noisy labeled datasets does not memorize noisy labels under a high learning rate. Thus, the noisy label of a sample can be corrected by reassigning the probability output of the DNN to the sample as a new label. We utilize this property to increase the number of ID samples during the training by cleaning the OOD scores of unlabeled data. To achieve this, the network parameters of the DNN and the OOD scores of unlabeled samples are alternately updated for each epoch and the OOD scores of unlabeled samples are reassigned to the estimation of OOD scores by DNN. 

At the same time, to achieve high performance on the classification of ID samples, we select ID samples in unlabeled data having low OOD scores and combine them with labeled ID samples for training the DNN to classify ID samples correctly in a semi-supervised manner. Although our method can be applied to any SSL method, we choose the state-of-the-art SSL method, MixMatch \cite{berthelot2019mixmatch}, in this paper.

\begin{figure}[t]
    \centering
    \includegraphics[width=0.8\textwidth]{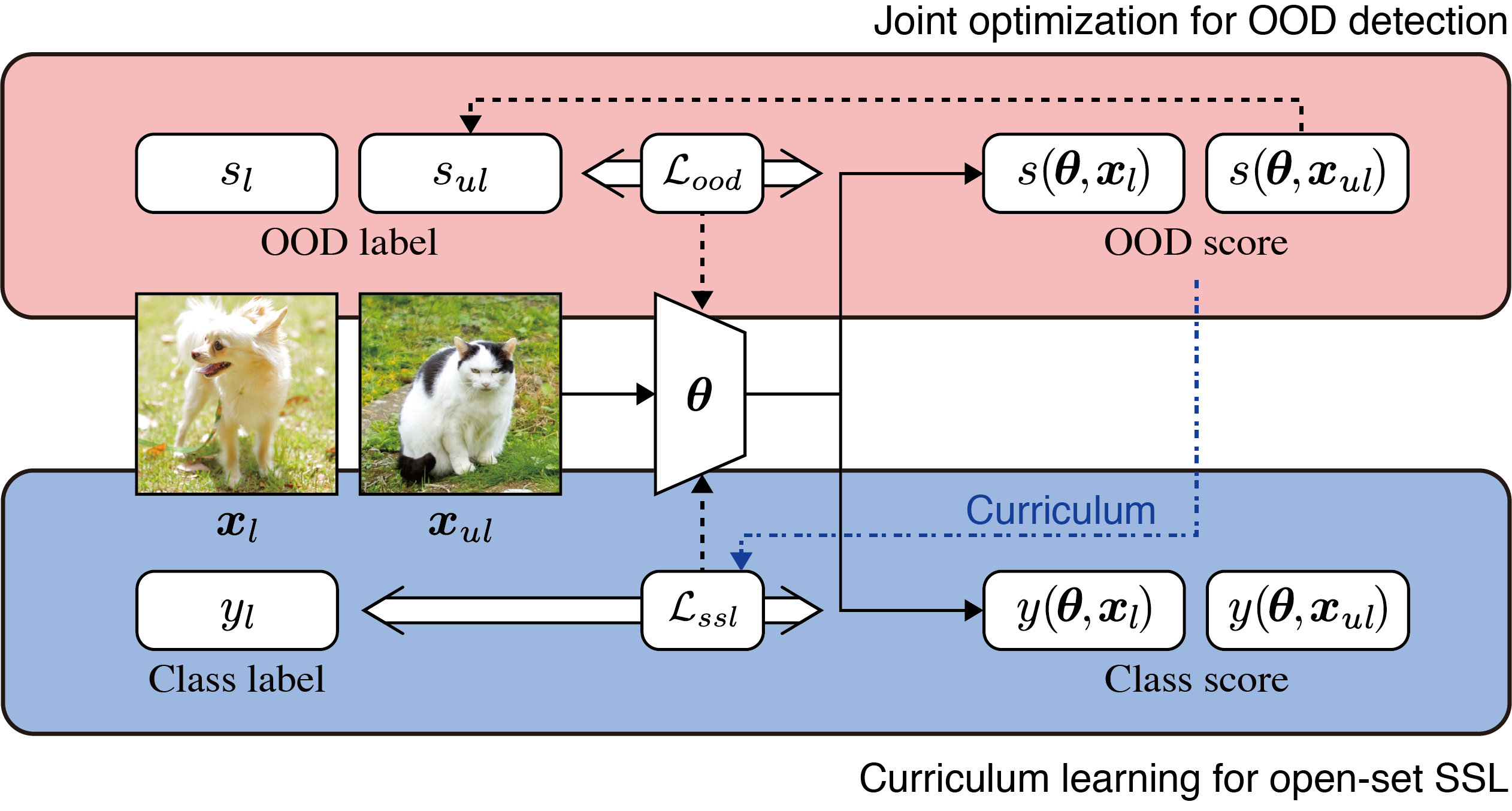}
    \caption{An overview of our framework. Noisy OOD scores of unlabeled data are reassigned to the outputs of OOD scores by the DNN. The network parameters and OOD scores are alternately updated for each epoch. The unlabeled samples used to calculate semi-supervised loss are selected by their OOD scores.}
    \label{fig:proposed}
\end{figure}

Instead of training OOD detection and ID classification separately, we further propose a multi-task training framework combining all the steps to formulate an end-to-end trainable network that can detect OOD samples and classify ID samples simultaneously. The overview of our framework is shown in \Fref{fig:proposed}.
  
\subsection{Training Procedure}
\subsubsection{Noisy Label Optimization for OOD Detection}

First, we assign an initial OOD score $s$ to all the data, which makes an ID image-label pair, $\{\boldsymbol{x}_{l},y_{l}\} \in \{X_{l}, Y_{l}\}$ become $\{\boldsymbol{x}_{l},y_{l},s_{l}\} \in \{X_{l}, Y_{l}, S_{l}\}$ , and an unlabeled image, $\boldsymbol{x}_{ul} \in X_{ul}$ become $\{\boldsymbol{x}_{ul},s_{ul}\} \in \{X_{ul}, S_{ul}\}$.
We initialize the OOD scores $S_{l}$ as 0 for the labeled (ID) samples, whereas the unlabeled samples are assigned the OOD score $S_{ul}$ of 1. This is denoted by:
\begin{equation}
    s_{l} = 0 \: (\forall s_{l} \in S_{l}),
\end{equation}
\begin{equation}
    s_{ul} = 1 \: (\forall s_{ul} \in S_{ul}).
\end{equation}

As a binary classification of OOD, parameters of the network $\boldsymbol{\theta}$ can be optimized as follows:
\begin{equation}
\min_{\boldsymbol{\theta}}\mathcal{L}_{ood}(\boldsymbol{\theta}|X_{l}, S_{l}, X_{ul}, S_{ul}),
\end{equation}
\begin{align}
    \mathcal{L}_{ood} = -{\frac{1}{|X_{l}|}}\sum_{i=1}^{|X_{l}|}(s_{l_i}\log s({\boldsymbol{\theta}},{\boldsymbol {x}_{l_i}})+(1-s_{l_i})\log (1-s({\boldsymbol{\theta}},{\boldsymbol {x}_{l_i}}))) \nonumber \\
    -{\frac{1}{|X_{ul}|}}\sum_{i=1}^{|X_{ul}|}(s_{ul_i}\log s({\boldsymbol{\theta}},{\boldsymbol {x}_{ul_i}})+(1-s_{ul_i})\log (1-s({\boldsymbol{\theta}},{\boldsymbol {x}_{ul_i}}))),
    \label{eq:crossentropy}
  \end{align}
  where $\mathcal{L}_{ood}$ denotes the cross entropy loss under the supervision of OOD score and $s(\boldsymbol{\theta},\boldsymbol {x}_{l_i})$ (or $s(\boldsymbol{\theta},\boldsymbol {x}_{ul_i})$) denotes the predicted OOD score of the image $\boldsymbol {x}_{l_i}$ (or $\boldsymbol {x}_{ul_i}$) with the network parameters being $\boldsymbol{\theta}$.

However, although the OOD scores $S_{ul}$ are initialized to 1 for all unlabeled samples, the existence of some ID samples in the unlabeled data leads to the classification of OOD as a noisy label problem. 
Tanaka et al. \cite{tanaka2018joint} showed that a network trained with a high learning rate is less likely to overfit to noisy labels, which means the loss \Eref{eq:crossentropy} is high for noisy labels and low for clean labels. So we obtain clean OOD scores by updating the OOD scores in the direction to decrease \Eref{eq:crossentropy}. Hence, we formulate the problem as the joint optimization of the network parameters and OOD scores as follows:
\begin{equation}
    \min_{\boldsymbol{\theta}, S_{ul}}\mathcal{L}_{ood}(\boldsymbol{\theta}, S_{ul}|X_{l}, S_{l}, X_{ul}),
\end{equation}

\begin{algorithm}[t]
    \caption{Joint Optimization}
    \label{alg:ao}
    \begin{algorithmic}
        \FOR{$t\leftarrow 0$ to $E$}
        \STATE $\text{update }\boldsymbol{\theta}^{(t+1)}\text{ by Adam on } \mathcal{L}_{ood}(\boldsymbol{\theta}^{t}, S_{ul}^{t}|X_{l}, S_{l}, X_{ul})$
        \STATE $\text{update }S_{ul}^{(t+1)}\text{ by \Eref{eq:soft}}$
        \ENDFOR
    \end{algorithmic}
\end{algorithm}
  
Alternately updating the network parameters $\boldsymbol{\theta}$ and OOD scores of unlabeled data $S_{ul}$ is achieved via joint optimization \cite{tanaka2018joint} by repeating the following two steps:

\textit{Updating $\boldsymbol{\theta}$ with fixed $S_{ul}$:} Since all terms in the loss function \Eref{eq:crossentropy} are differentiable with respect to $\boldsymbol{\theta}$, we update $\boldsymbol{\theta}$ by the Adam optimizer \cite{kingma2014adam} on \Eref{eq:crossentropy}.
  
\textit{Updating $S_{ul}$ with fixed $\boldsymbol{\theta}$:} Considering the update of $S_{ul}$,  we need to minimize $\mathcal{L}_{ood}$ with fixed $\boldsymbol{\theta}$ to correct OOD scores. $\mathcal{L}_{ood}$ can be minimized when the predicted OOD scores of the network equals $S_{ul}$. As a result, $S_{ul}$ is updated as follows: 
  \begin{equation}
  \label{eq:soft}
  s_{ul_i} \leftarrow s(\boldsymbol{\theta},\boldsymbol {x}_{ul_i}).
  \end{equation}
  
The whole algorithm of joint optimization is shown in \Aref{alg:ao}.

\subsubsection{Multi-task Curriculum Learning for Open-set SSL}
As general SSL, the optimization problem of the network parameters $\boldsymbol{\theta}$ is formulated as follows:
\begin{equation}
\min_{\boldsymbol{\theta}}\mathcal{L}_{ssl}(\boldsymbol{\theta}|X_{l}, Y_{l}, X_{ul}),
\end{equation}
where $\mathcal{L}_{ssl}$ denotes a loss function such as the sum of cross-entropy loss on labeled data and L2 loss on unlabeled data in \cite{laine2016temporal}.

During the training of the network for OOD detection, we also train the network to classify ID data by SSL. As a result, our method is a multi-task learning problem formulated as follows:
\begin{equation}
    \label{eq:all}
    \min_{\boldsymbol{\theta}, S_{ul}}\mathcal{L}_{ssl}(\boldsymbol{\theta}|X_{l}, Y_{l}, X_{ul}) + \mathcal{L}_{ood}(\boldsymbol{\theta}, S_{ul}|X_{l}, S_{l}, X_{ul}).
\end{equation}

The inclusion of the training for ID data classification is helpful to OOD detection because the network can learn more discriminative features. However, while optimizing $\boldsymbol{\theta}$ on the semi-supervised part $\mathcal{L}_{ssl}(\boldsymbol{\theta}^{t}|X_{l}, Y_{l}, X_{ul})$ in \Eref{eq:all}, the existence of OOD samples in $X_{ul}$ is detrimental to the training for the classification of ID samples. This problem is solved by using curriculum learning that picks up ID samples $X_{ul}^{id}$ from unlabeled data $X_{ul}$ according to the OOD scores of unlabeled data $S_{ul}$. 

Although we can simply sample top $\eta\%$ samples from $X_{ul}$ in ascending order of OOD scores $S_{ul}$, we implement Otsu thresholding \cite{otsu1979threshold} to decide the threshold $th_{otsu}$ automatically, which reduces one hyper-parameter. Then, the selected unlabeled samples are denoted as follows:
\begin{equation}
    \label{eq:curr}
    X_{ul}^{id}=\{\boldsymbol {x}_{ul_i}|s(\boldsymbol{\theta},\boldsymbol {x}_{ul_i})<th_{otsu}, 1\le i \le N \},
\end{equation}
which converts the total loss function to:
\begin{equation}
    \min_{\boldsymbol{\theta}, S_{ul}}\mathcal{L}_{ssl}(\boldsymbol{\theta}|X_{l}, Y_{l}, X_{ul}^{id}) + \mathcal{L}_{ood}(\boldsymbol{\theta}, S_{ul}|X_{l}, S_{l}, X_{ul}).
\end{equation}

\begin{algorithm}[t]
    \caption{Multi-task curriculum learning}
    \label{alg:pro}
    \begin{algorithmic}
        \FOR{$t\leftarrow 0$ to $E$}
        \STATE Select ID samples in unlabeled data $X_{ul}^{id}$ by \Eref{eq:curr}
        \STATE $\text{update }\boldsymbol{\theta}^{(t+1)}\text{ by Adam on }\mathcal{L}_{ssl}(\boldsymbol{\theta}^{t}|X_{l}, Y_{l}, X_{ul}^{id}) + \mathcal{L}_{ood}(\boldsymbol{\theta}^{t}, S_{ul}^{t}|X_{l}, S_{l}, X_{ul})$
        \STATE $\text{update }S_{ul}^{(t+1)}\text{ by \Eref{eq:soft}}$
        \ENDFOR
    \end{algorithmic}
\end{algorithm}

The entire algorithm of our method is as shown in \Aref{alg:pro}. It is to be noted that the semi-supervised loss $\mathcal{L}_{ssl}(\boldsymbol{\theta}^{t}|X_{l}, Y_{l}, X_{ul}')$ is a general loss of SSL, which implies that our method is applicable to any SSL method. In this paper, we use MixMatch \cite{berthelot2019mixmatch} as SSL.

\section{Experiments}
\label{sec:exp}
In this section, we discuss our experimental settings and results. We demonstrate the effectiveness of our method on a diverse set of in- and out-of-distribution dataset pairs for open-set SSL. We found that our method outperformed the current state-of-the-art methods by a considerable margin. We used PyTorch 1.1.0 \cite{paszke2017automatic} to run all the experiments.

\subsection{Neural Network Architecture}
\label{sec:nn}
Following \cite{berthelot2019mixmatch}\cite{oliver2018realistic}, we implemented our network based on Wide ResNet (WRN) \cite{Zagoruyko2016WRN}. We first trained the model only on the OOD classification loss for 100 epochs and the update of the OOD scores was set to start at the 10th epoch, to achieve stable performance. The model was then trained on the total loss function in \Aref{alg:pro} for 1,024 epochs and 1,024 iterations of each epoch, which is the same as \cite{berthelot2019mixmatch}. We used the Adam \cite{kingma2014adam} optimizer and the learning rate was set as 0.002. 64 samples each from the labeled and unlabeled data are sampled for a batch. We report the average test accuracy of the last 10 checkpoints.
    
\subsection{In-Distribution Datasets}
CIFAR-10 \cite{Krizhevsky09learningmultiple} and SVHN \cite{netzer2011reading} (each containing 10 classes) were used as in-distribution datasets. A total of 5,000 samples were split from the original training data as validation data, and all original test samples were used for testing. We further split the remaining training samples (45,000 for CIFAR-10 and 68,257 for SVHN) into labeled and unlabeled data. Following \cite{oliver2018realistic}\cite{berthelot2019mixmatch}, we used \{250, 1000, 4000\} samples as labeled data and remaining samples as unlabeled data.

\subsection{Out-of-Distribution Datasets}
\begin{figure}[t]
    \centering
    \includegraphics[width=0.6\textwidth]{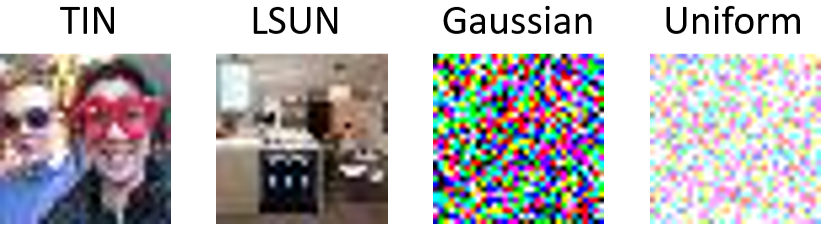}
    \caption{Examples of Out-of-Distribution Datasets. The samples from TinyImageNet and LSUN are resized to $32 \times 32$.}
    \label{fig:ood}
\end{figure}

As the OOD data are mixed in the unlabeled data, we added 10,000 samples from the following four datasets for each setting:
\begin{enumerate}
    \item \textbf{TinyImageNet (TIN).} The Tiny ImageNet dataset \cite{imagenet_cvpr09} contains 10,000 test images from 200 different classes, which are drawn from the original 1,000 classes of ImageNet \cite{imagenet_cvpr09}. All samples are downsampled by resizing the original image to a size of $32 \times 32$.
    \item \textbf{LSUN.}  The Large-scale Scene Understanding dataset (LSUN) consists of 10,000 test images from 10 different scene categories.\cite{song2015construction}. Similar to TinyImageNet, all samples are downsampled by resizing the original image to a size of $32 \times 32$.
    \item \textbf{Gaussian.} The synthetic Gaussian noise dataset contains 10,000 random 2D Gaussian noise images, where each RGB value of every pixel is sampled from an independent and identically distributed Gaussian distribution with mean 0.5 and unit variance. We further clip each pixel value into the range $[0, 1]$.
    \item \textbf{Uniform.} The synthetic uniform noise dataset contains 10,000 images where each RGB value of every pixel is independently and identically sampled from a uniform distribution on $[0, 1]$.
\end{enumerate}
Examples of these datasets are shown in \Fref{fig:ood}.
The experimental setting is summarized in \Tref{tbl:exp}.

\begin{table*}[t]
    \centering
    \caption{The number and type of labeled and unlabeled samples in the experimental setting.}
    \tabcolsep = 1.1mm
    \scalebox{0.9}{
    \begin{tabular}{c|c|c|c|c}
    \multicolumn{5}{c}{\textbf{CIFAR-10}}                                                                                                       \\\toprule
    \#Labeled & \#Unlabeled & \begin{tabular}[c]{@{}c@{}}\#Unlabeled\\ Outlier\end{tabular} & \#Valid                 & \#Test                   \\\hline\hline
    250     & 44,750     & \multirow{3}{*}{10,000}                                      & \multirow{3}{*}{5,000} & \multirow{3}{*}{10,000} \\\cline{1-2}
    1,000    & 44,000     &                                                             &                       &                        \\\cline{1-2}
    4,000    & 41,000     &                                                             &                       &                        \\\bottomrule
    \multicolumn{5}{c}{}\\
    \multicolumn{5}{c}{\textbf{SVHN}}                                                                                                           \\\toprule
    \#Labeled & \#Unlabeled & \begin{tabular}[c]{@{}c@{}}\#Unlabeled\\ Outlier\end{tabular} & \#Valid                 & \#Test                   \\\hline\hline
    250     & 68,007     & \multirow{3}{*}{10,000}                                      & \multirow{3}{*}{5,000} & \multirow{3}{*}{26,032} \\\cline{1-2}
    1,000    & 67,257     &                                                             &                       &                        \\\cline{1-2}
    4,000    & 64,257     &                                                             &                       &                        \\\bottomrule
    \end{tabular}
    }
    \label{tbl:exp}
    \end{table*}

\begin{table*}[t]
    \centering
    \caption{Accuracy (\%) for CIFAR-10/SVHN and OOD dataset pairs. We report the averages and the standard deviations of the scores obtained from three trials. Bold values represent the highest accuracy in each setting. \textit{Clean} shows the upper limit of the model when the unlabeled data contains no OOD data.}
    \label{tbl:res}
    \tabcolsep = 1.1mm
    \scalebox{0.85}{
    \begin{tabular}{c|cc|cc|cc}
    \multicolumn{7}{c}{\textbf{CIFAR-10}} \\
    \toprule
    \multirow{2}{*}{\begin{tabular}{c}OOD \\ dataset\end{tabular}}    & \multicolumn{2}{c|}{250 labeled} & \multicolumn{2}{c|}{1000 labeled} & \multicolumn{2}{c}{4000 labeled} \\\cline{2-7}
              & Baseline  & Ours  & Baseline   & Ours  & Baseline   & Ours  \\\hline\hline
    TIN      & 82.42 $\pm$ 0.70 & \textbf{86.44 $\pm$ 0.64} & 88.03 $\pm$ 0.22 &       \textbf{89.85 $\pm$ 0.11} & 91.25 $\pm$ 0.13 & \textbf{93.03 $\pm$ 0.05} \\
    LSUN     & 76.32 $\pm$ 4.19 & \textbf{86.65 $\pm$ 0.41} & 87.03 $\pm$ 0.41 &       \textbf{90.19 $\pm$ 0.47} & 91.18 $\pm$ 0.33 & \textbf{92.91 $\pm$ 0.03} \\
    Gaussian & 75.76 $\pm$ 3.49 & \textbf{87.34 $\pm$ 0.13} & 85.71 $\pm$ 1.14 &       \textbf{89.80 $\pm$ 0.26} & 91.51 $\pm$ 0.35 & \textbf{92.53 $\pm$ 0.08} \\
    Uniform  & 72.90 $\pm$ 0.96 & \textbf{85.54 $\pm$ 0.11} & 84.49 $\pm$ 1.06 &       \textbf{89.87 $\pm$ 0.08} & 90.47 $\pm$ 0.38 & \textbf{92.83 $\pm$ 0.04} \\
    \hline\hline 
    Clean & \multicolumn{2}{c|}{87.65 $\pm$ 0.29} & \multicolumn{2}{c|}{90.67 $\pm$ 0.29} & \multicolumn{2}{c}{93.30 $\pm$ 0.10}\\
    \bottomrule
    \multicolumn{7}{c}{} \\
    \multicolumn{7}{c}{\textbf{SVHN}} \\
    \toprule
    \multirow{2}{*}{\begin{tabular}{c}OOD \\ dataset\end{tabular}}    & \multicolumn{2}{c|}{250 labeled} & \multicolumn{2}{c|}{1000 labeled} & \multicolumn{2}{c}{4000 labeled} \\\cline{2-7}
              & Baseline  & Ours  & Baseline   & Ours  & Baseline   & Ours  \\\hline\hline
    TIN      & 94.66 $\pm$ 0.14 & \textbf{95.21 $\pm$ 0.27} & 95.58 $\pm$ 0.38 &       \textbf{96.65 $\pm$ 0.14} & 96.73 $\pm$ 0.05 & \textbf{97.01 $\pm$ 0.03} \\
    LSUN     & 94.98 $\pm$ 0.23 & \textbf{95.40 $\pm$ 0.17} & 95.46 $\pm$ 0.05 &       \textbf{96.51 $\pm$ 0.16} & 96.75 $\pm$ 0.01 & \textbf{97.15 $\pm$ 0.02} \\
    Gaussian & 93.42 $\pm$ 1.09 & \textbf{95.23 $\pm$ 0.04} & 95.85 $\pm$ 0.33 &       \textbf{96.50 $\pm$ 0.11} & 96.97 $\pm$ 0.02 & \textbf{97.07 $\pm$ 0.07} \\
    Uniform  & 94.78 $\pm$ 0.25 & \textbf{95.07 $\pm$ 0.12} & 95.62 $\pm$ 0.50 &       \textbf{96.47 $\pm$ 0.24} & 96.86 $\pm$ 0.12 & \textbf{97.04 $\pm$ 0.02} \\
    \hline\hline 
    Clean & \multicolumn{2}{c|}{96.04 $\pm$ 0.39} & \multicolumn{2}{c|}{96.84 $\pm$ 0.06} & \multicolumn{2}{c}{97.23 $\pm$ 0.05}\\
    \bottomrule
    \end{tabular}
    }
\end{table*}

\begin{figure}[t]
    \centering
    \includegraphics[width=0.95\textwidth]{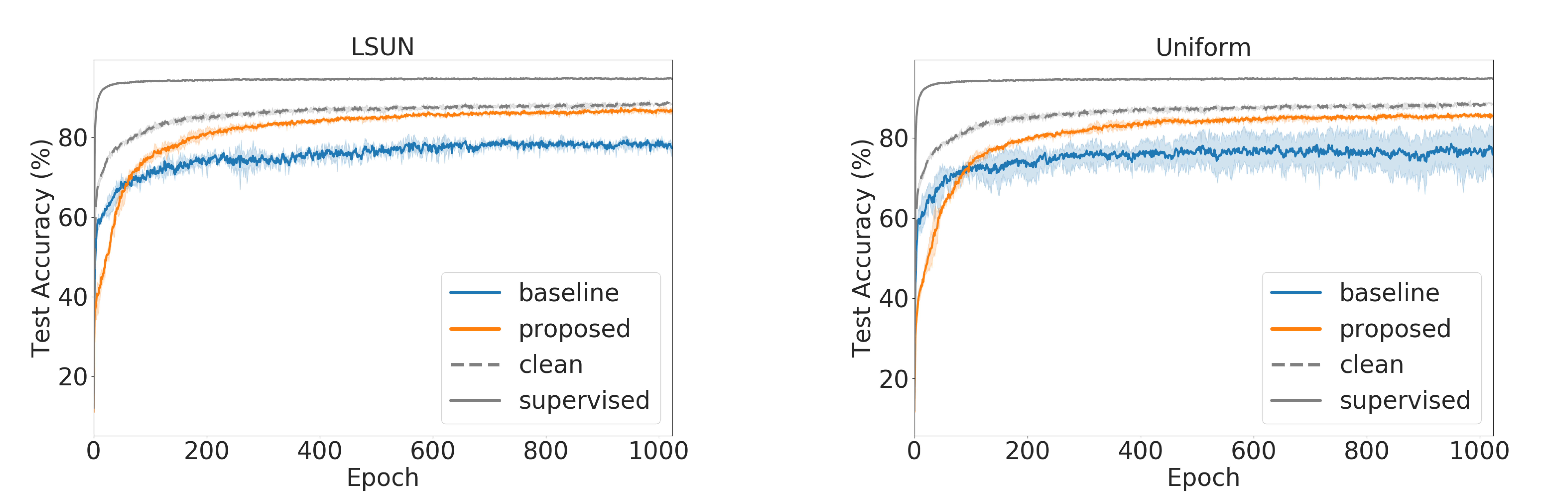} 
    \caption{Test accuracy vs. the number of epochs using CIFAR-10 as ID and other datasets as OOD when 250 labeled samples are used. \textit{Clean} shows the upper limit of the model when the unlabeled data contains no OOD data and \textit{supervised} shows the performance when all the samples are labeled and no OOD data is contained.}
    \label{fig:acc}
\end{figure}

\begin{table}[t]
    \centering
    \caption{Performance (\%) of OOD detection by our proposed method. We report the averages of the scores obtained from three trials.}
    \label{tbl:prec_cifar}
    \tabcolsep = 1.1mm
    \scalebox{0.9}{
    \begin{tabular}{c|cc|cc|cc}
    \multicolumn{7}{c}{\textbf{CIFAR-10}} \\
    \toprule
    \multirow{2}{*}{\begin{tabular}{c}OOD \\ dataset\end{tabular}}    & \multicolumn{2}{c|}{250 labeled} & \multicolumn{2}{c|}{1000 labeled} & \multicolumn{2}{c}{4000 labeled} \\\cline{2-7}
                & Recall  & Precision  & Recall   & Precision  & Recall   & Precision  \\\hline\hline
    TIN      & 99.22 & 98.48 & 99.48 & 97.11  & 100.00 & 99.30 \\
    LSUN     & 99.48 & 99.38 & 99.95 & 98.95 & 100.00 & 99.64 \\
    Gaussian & 100.00 & 100.00 & 100.00 & 100.00  & 100.00 & 100.00 \\
    Uniform  & 100.00 & 100.00 & 100.00 & 100.00  & 100.00 & 100.00 \\
    \bottomrule
    \multicolumn{7}{c}{} \\
    \multicolumn{7}{c}{\textbf{SVHN}} \\
    \toprule
    \multirow{2}{*}{\begin{tabular}{c}OOD \\ dataset\end{tabular}}    & \multicolumn{2}{c|}{250 labeled} & \multicolumn{2}{c|}{1000 labeled} & \multicolumn{2}{c}{4000 labeled} \\\cline{2-7}
        & Recall  & Precision  & Recall   & Precision  & Recall   & Precision  \\\hline\hline
        TIN      & 84.28 & 99.93 & 98.4 & 99.83  & 99.59 & 99.87  \\
        LSUN     & 88.55  & 99.98 & 98.28 & 99.97 & 99.70  & 99.98 \\
        Gaussian & 87.52 & 100.00 & 99.28 & 100.00  & 99.76 & 100.00 \\
        Uniform  & 82.67 & 100.00 & 99.21 & 100.00  & 99.78 & 100.00 \\
    \bottomrule
    \end{tabular}
    }
\end{table}

\subsection{Results}
The results for the CIFAR-10 dataset are summarized in the upper part of \Tref{tbl:res}, which shows the comparison of our method and the baseline. We used the original MixMatch \cite{berthelot2019mixmatch} without any OOD detection as the baseline method. \Tref{tbl:res} clearly shows that our approach significantly outperforms the baseline by eliminating the effect of OOD samples in unlabeled data. Compared to TIN and LSUN, which are natural images, synthetic datasets (Gaussian and Uniform) are more harmful to the performance of SSL. Our technique has successfully enabled SSL methods to achieve stable performance on these outliers by detecting them. 

In \Fref{fig:acc}, we show test accuracy vs. the number of epochs. We observe that our method continuously improves the performance of the model during the latter half of the training process and its performance is more stable compared to the baseline method during the training. At the beginning of the training process, our method is observed to converge slower than the baseline method. We consider this to be due to the multi-task learning, where our method also learns to detect OOD samples.

The lower part of \Tref{tbl:res} shows the comparison of our method and the baseline for the SVHN dataset. Compared to the case where CIFAR-10 is used as the ID dataset, the effect of outliers on the model is much smaller in this case, possibly because the classification of SVHN is a comparatively easier task. In this situation, our method continues to exhibit a higher and more stable performance than the baseline.

We also studied the performance of OOD detection by the proposed method. Since we use the threshold calculated by Otsu thresholding to select ID samples in unlabeled data, we evaluate the performance of OOD detection by precision and recall. Precision is calculated by the percentage of ID samples in the selected samples by curriculum learning. Recall is calculated by the percentage of selected ID samples among all ID samples in unlabeled data. Higher precision and recall indicate better OOD detection -- the precision and recall of a perfect detector are both 1. \Tref{tbl:prec_cifar} shows the results. We find that our method achieves high precision and recall in all the cases, indicating that our method successfully in selecting ID samples from unlabeled data for semi-supervised training.

\subsection{Ablation Studies}
We further analyzed the effects of the following factors:

\textbf{The number of OOD samples in the unlabeled data.} We used LSUN as OOD and we changed the number of OOD samples in $X_{ul}$.
The result is summarized in \Tref{tbl:res_cifar_ab2}, which shows that our proposed method works under different OOD conditions except for a case with few outliers in unlabeled data. 
When there are more OOD samples in the unlabeled data, the performance of the baseline model is lower while our method can achieve stable performance.

\begin{table}[t]
    \centering
    \caption{Accuracy (\%) for CIFAR-10 as ID and LSUN as OOD on different numbers of OOD samples when 250 labeled samples are used.}
    \label{tbl:res_cifar_ab2}
    \tabcolsep = 1.1mm
    \scalebox{0.9}{
        \begin{tabular}{c|cc|cc|cc|cc}
            \toprule
            \multirow{2}{*}{\#OOD samples}    & \multicolumn{2}{c|}{2000} & \multicolumn{2}{c|}{5000} & \multicolumn{2}{c|}{10000} & \multicolumn{2}{c}{20000}\\\cline{2-9}
                  & Baseline  & Ours  & Baseline   & Ours  & Baseline   & Ours  & Baseline   & Ours \\\hline\hline
        Accuracy (\%)    & \textbf{88.16} & 84.82 & 82.98 & \textbf{86.20} & 76.32 & \textbf{86.65} & 70.3 & \textbf{85.83}
        \\ \bottomrule
        \end{tabular}
    }
\end{table}

\textbf{The performance of OOD detection compared to existing OOD detection methods.} As mentioned in \Sref{sec:related}, the existing OOD detection methods cannot achieve high performance when the labeled data is limited and the unlabeled data cannot be utilized in training. We show the results of applying the existing OOD detection method in the setting of open-set semi-supervised learning in \Tref{tbl:res_cifar_ood}. We choose the most challenging cases when only 250 labeled samples are available. We report the AUROC (the area under the false positive rate against the true positive rate curve) of the OOD detection score of each method. The AUROC of a perfect detector is 1. \Tref{tbl:res_cifar_ood} shows that our approach significantly outperforms other OOD detection methods. It is also interesting that the AUROC of previous methods is less than 50\% in some cases, which means the model shows more confidence on the predictions of OOD samples than those of ID samples (and this observation conflicts with the idea of \cite{hendrycks2016baseline}\cite{liang2017enhancing}).

\begin{table}[t]
    \centering
    \caption{The comparison of AUROC (\%) in the task of OOD Detection.}
    \label{tbl:res_cifar_ood}
    \tabcolsep = 1.1mm
    \scalebox{0.9}{
        \begin{tabular}{C{2cm}|C{2cm}|C{2cm}|C{2cm}|C{2cm}}
            \toprule
        {\begin{tabular}{c}ID \\ dataset\end{tabular}}&{\begin{tabular}{c}OOD \\ dataset\end{tabular}} & {\begin{tabular}{c}Hendrycks \\\& Gimpel \cite{hendrycks2016baseline}\end{tabular}}  & ODIN \cite{liang2017enhancing}  & Ours \\\hline\hline
            \multirow{4}{*}{CIFAR-10}&TIN  & 50.92 & 54.54 & \textbf{98.86}\\
            &LSUN  & 54.34 & 58.02 & \textbf{99.82}\\
            &Gaussian  & 32.41 & 37.49 & \textbf{100.00}\\
            &Uniform  & 45.43 & 51.05 & \textbf{100.00}\\\midrule
            \multirow{4}{*}{SVHN}&TIN  & 50.48 & 57.09 & \textbf{99.57}\\
            &LSUN  & 51.44 & 53.68 & \textbf{99.84}\\
            &Gaussian  & 21.20 & 1.87 & \textbf{99.98}\\
            &Uniform  & 2.79 & 8.31 & \textbf{99.97}
        \\ \bottomrule
        \end{tabular}
    }
  \end{table}

\begin{figure}[t]
    \centering
    \includegraphics[width=0.8\textwidth]{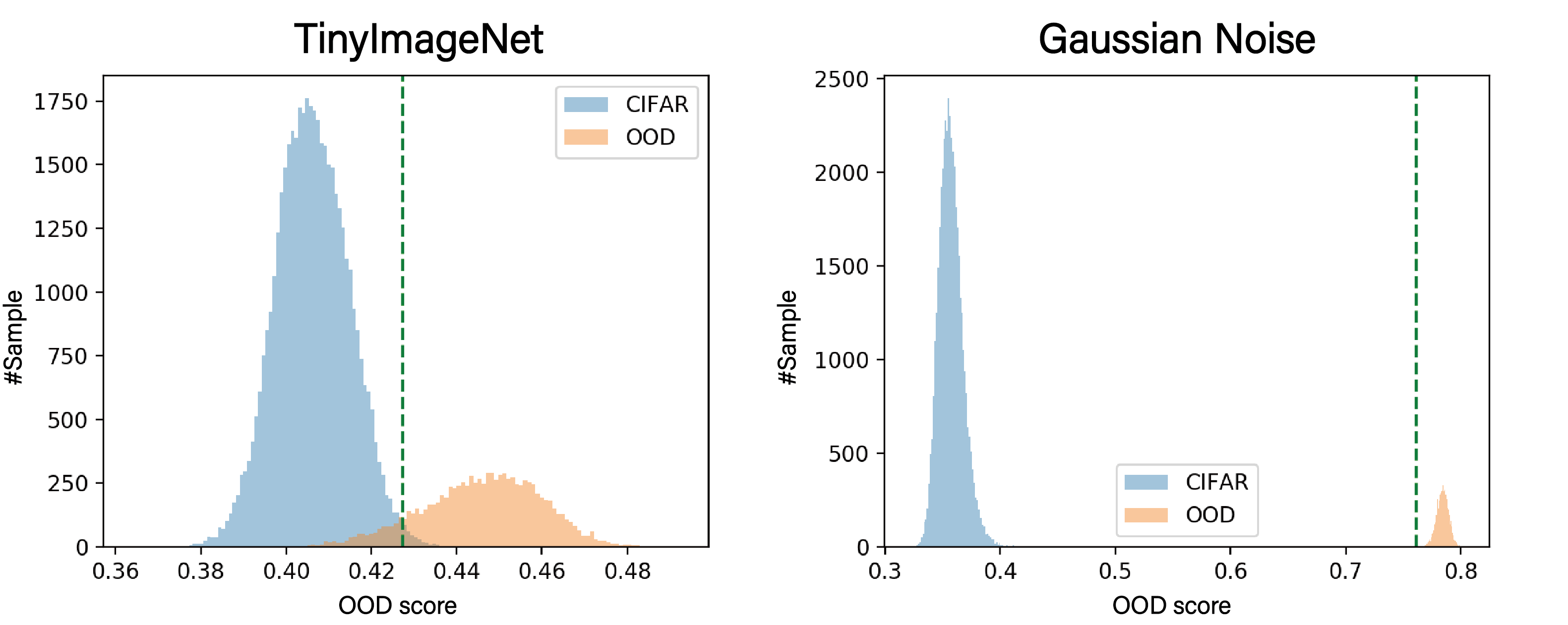}
    \caption{The histogram of OOD scores and the threshold (the green line) calculated by Otsu thresholding  when 250 labeled samples are used.}
    \label{fig:hist}
\end{figure}

\textbf{The performance of Otsu thresholding.} We use Otsu thresholding to calculate the threshold for splitting ID and OOD samples for the curriculum learning of semi-supervised learning. \Fref{fig:hist} shows the histogram of OOD scores and the threshold calculated using Otsu thresholding. We find that both ID and OOD samples can be successfully separated by the threshold. 

\subsection{Discussion}
\textbf{Limitations.} As shown in \Tref{tbl:res_cifar_ab2}, our method fails to improve the baseline if there are few outliers in the unlabeled data. This failure mainly comes from the wrong threshold calculated by Otsu thresholding, since the number of ID samples and OOD samples is extremely imbalanced. This problem can be solved by changing the OOD threshold, which means we can introduce a new parameter to control the number of unlabeled samples selected as ID data.

\textbf{``Similar'' outliers.} The outlier datasets used in \Sref{sec:exp} are still quite different from the original training datasets CIFAR-10 and SVHN. Including similar outliers in the unlabeled data is a more complicated scenario. We tried using the animal classes in CIFAR-10 as ID and other classes as OOD and found that these similar outliers are not as harmful as dissimilar outliers, which leads to a 3\% decrease in test accuracy. Our method can still reach 2\% higher than the test accuracy of the baseline with 94\% precision and 98\% recall of OOD detection.

\textbf{Additional comparisons.} Chen et al. \cite{chen2020semi} works on a close setting to our paper at around the same time as our paper, but there are two significant differences between \cite{chen2020semi} and our work. First, the experimental setting is different. \cite{chen2020semi} defines the mismatch of class distribution as some known classes are not contained in the unlabeled dataset and some unknown classes are contained in the unlabeled dataset. Second, the method of utilizing OOD samples in unlabeled data is different. In \cite{chen2020semi}, the OOD samples are simply ignored by filtering the confidence score, which means they mainly train an SSL model and just use the output of the model directly.

For the comparison, we implemented \cite{chen2020semi} for filtering OOD in MixMatch and show the comparison in the setting of CIFAR-10 with 250 labeled samples. The results are summarized in \Tref{tbl:add_res} and it shows our method has better performance not only in SSL but also in OOD detection, because we explicitly train an OOD detector by unlabeled OOD samples together with the training of SSL model, which enables our method to achieve higher OOD detection performance.

\begin{table}[t]
    \centering
    \caption{Performance (\%) of UASD \cite{chen2020semi} and our propsoed method.}
    \label{tbl:add_res}
    \tabcolsep = 1.1mm
    \scalebox{0.9}{
        \begin{tabular}{c|ccc|cc|cc}
            \toprule
            \multirow{2}{*}{\begin{tabular}{c}OOD \\ dataset\end{tabular}}    & \multicolumn{3}{c|}{Test Accuracy} & \multicolumn{2}{c|}{Detection Recall} & \multicolumn{2}{c}{Detection Precision} \\\cline{2-8}
                        & Baseline  & UASD \cite{chen2020semi}  & Ours  & UASD \cite{chen2020semi} & Ours  & UASD \cite{chen2020semi}  & Ours  \\\hline\hline
            TIN      & 82.42 & 83.53 & \textbf{86.44} & 66.47  & \textbf{99.22} & 96.84 & \textbf{98.48} \\
            LSUN     & 76.32 & 80.87 & \textbf{86.65} & 63.88 & \textbf{99.48} & 96.50  & \textbf{99.38}
            \\
            \bottomrule
            \end{tabular}
    }
  \end{table}

\section{Conclusion}
In this paper, we have proposed a multi-task curriculum for open-set SSL, where the labeled data is limited and the unlabeled data contains some OOD samples. To detect these OOD samples, our method utilizes joint optimization framework to estimate the probability of the unlabeled sample belonging to OOD, which is achieved by updating the network parameters and the OOD score alternately. Simultaneously, we use curriculum learning to exclude these OOD samples from semi-supervised learning and utilize these data with labeled data for training the model to classify ID samples with high performance. We evaluated our method on several open-set semi-supervised benchmarks and proved that our method achieves state-of-the-art performance by detecting the OOD samples with high accuracy.

\subsubsection*{ACKNOWLEDGEMENTS}
This work was supported by JSPS KAKENHI Grant Number 18H03254 and JST CREST Grant Number \mbox{JPMJCR1686}, Japan.
  
\clearpage
\bibliographystyle{splncs04}
\bibliography{egbib}
\end{document}